\documentclass{IEEEtran}
\usepackage{array}
\usepackage{graphicx}
\usepackage{amsmath}
\usepackage{amssymb}
\usepackage[numbers]{natbib}
\usepackage[T1]{fontenc}
\usepackage{lmodern}
\usepackage{booktabs}
\usepackage{url}
\usepackage{float}
\usepackage{placeins}
\usepackage{needspace}
\usepackage{algpseudocode}

\newcommand{\decisiontype}[1]{\textit{#1}}
\newcounter{rdpalgorithm}
\begin{document}
\nocite{ref5,ref6,ref7,ref8,ref9,ref10,ref11,ref12,ref13,ref14,ref15,ref16,ref17,ref19,ref20,ref21,mozannar2020defer,ref23,ref24,ref25,ref26,ref27,ref28,ref29,ref30,ref31,ref32,ref33,ref34,ref35,ref36,ref37,ref38,ref39,ref40,ref41,ref42,ref44,ref45}

\title{Decision-Centric Research: A Calibrated Decision Model and Benchmark for Scientific Agents}

\author{\IEEEauthorblockN{Shaoning Zeng\textsuperscript{1}, Wen Huang\textsuperscript{1}, Xinzhong Zhu, Yi Duan, Jianhang Zhou, Jianping Gou}}

\maketitle

\begin{abstract}

Scientific agents make consequential choices about evidence, next actions and research strategy, but choices embedded in generated text are difficult to evaluate independently. \textbf{Decision-Centric Research (DCR)} is introduced to make individual choices the common unit of supervision, prediction and evaluation through typed predictions over an explicit Research State. DCR defines 24 decision types, ten of which are instantiated in the \textbf{Research Decision Benchmark (RD-Bench)} of 2,000 rule-generated cases. The \textbf{Decision-Centric Research Model (DCRM)} shares a candidate-token readout across types to produce answers and option probabilities in one forward pass. Under a held-out research-domain split, a 0.6B DCRM (state-aware) achieves \textbf{89.4\% Top-1 accuracy}, compared with \textbf{27.8\%} for DCRM (question-only) trained with the same recipe and \textbf{36.4\%} for DeepSeek-V4.1-Flash. All 30 type--seed comparisons against DCRM (question-only) are significant after Holm correction. After held-out calibration, the resulting prediction sets contain a mean of \textbf{1.515 candidates at 97.1\% empirical label coverage}, reducing mean set size by \textbf{57.4\%} relative to DCRM (question-only) at similar coverage. These results establish a compact state-aware decision layer with accurate typed prediction and an inspectable uncertainty interface.

\end{abstract}

\begin{IEEEkeywords}
Scientific agents, research decision-making, compact language models, conformal prediction, decision benchmarks.
\end{IEEEkeywords}

\section{Introduction}
\label{sec:introduction}

Scientific agents bring language-model reasoning and tool use into workflows that require repeated choices about evidence, experiments and research direction. ReAct interleaved verbal reasoning with actions to support interaction with external environments \cite{ref5}. Cognitive Architectures for Language Agents (CoALA) organized memory, structured action spaces and decision making within a common framework \cite{ref8}. In scientific research, Agent Laboratory integrated literature review, experimentation and report writing into a workflow with human feedback \cite{ref34}. Across these workflows, choices about which evidence to accept, which experiment to run and when to revise a direction guide subsequent actions and resource use.

Evaluating these choices calls for a link between task performance and individual decisions. MLAgentBench evaluated agents on machine-learning experimentation tasks \cite{ref35}, while MLE-bench assessed machine-learning engineering through Kaggle competitions \cite{ref36}. Such task-level outcomes characterize overall performance, whereas an individual research choice requires its own candidate space and scoring target. Uncertainty methods provide a complementary foundation for inspecting the alternatives to a selected answer. Conformal prediction constructs prediction sets with coverage guarantees under exchangeability \cite{ref15}, and KnowNo applied this approach to candidate plans and human assistance in robot planning \cite{ref16}. For scientific agents, these lines of work motivate a common interface that connects research context, candidate-level evaluation and uncertainty over the available choices.

\begin{figure*}[t]
\centering
\includegraphics[width=\textwidth]{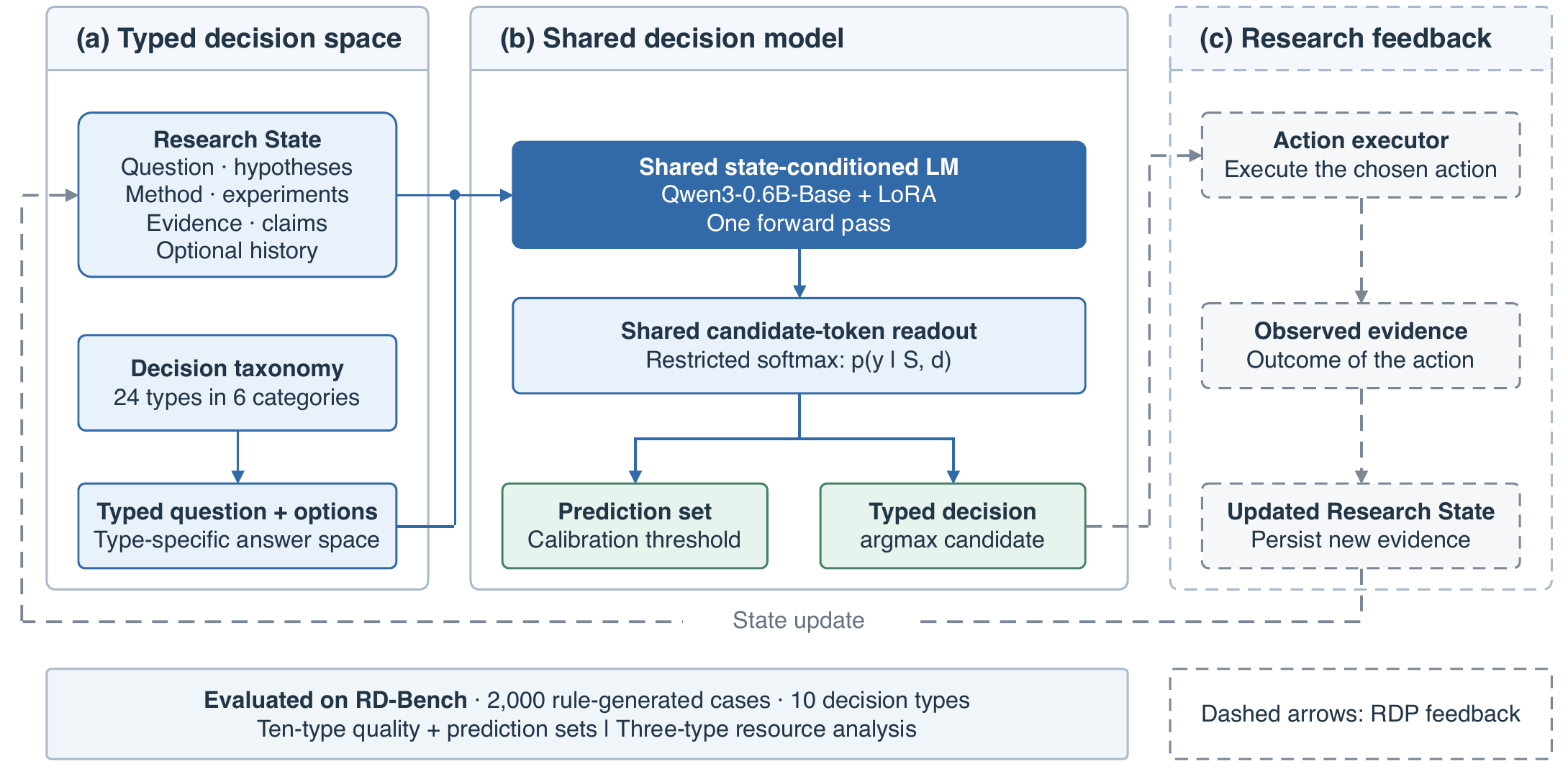}
\caption{Decision-Centric Research models scientific choices as state-conditioned, typed predictions. (a) Research State and the decision taxonomy define the context and candidate space. (b) A shared language model with Low-Rank Adaptation (LoRA) produces a candidate-token distribution in one forward pass, supporting both decision selection and calibrated prediction sets. (c) The Research Decision Process (RDP) connects decisions, actions, evidence and persistent state updates. Dashed arrows indicate the RDP feedback path.}
\label{fig:dcr-overview}
\end{figure*}

To evaluate individual research choices independently of generated text, we introduce \textbf{Decision-Centric Research (DCR)}, which represents each choice as a typed prediction conditioned on an explicit Research State. The \textbf{Decision-Centric Research Model (DCRM)} shares a candidate-token readout across types, producing an answer and option probabilities in one forward pass. Held-out calibration turns these probabilities into prediction sets. The \textbf{Research Decision Process (RDP)} links decisions to actions and evidence updates, as illustrated in Figure~\ref{fig:dcr-overview}.

The \textbf{Research Decision Benchmark (RD-Bench)} evaluates this interface on 2,000 rule-generated cases spanning ten decision types. Under a held-out research-domain split, the 0.6B DCRM (state-aware) achieves 89.4\% Top-1 accuracy, compared with 27.8\% for DCRM (question-only) using the same base model and training recipe. Its calibrated prediction sets contain a mean of 1.515 candidates at 97.1\% empirical label coverage. These findings support accurate typed prediction and compact prediction sets from a shared option distribution. The main contributions of this work are summarized as follows.

\begin{enumerate}
\item \textbf{A typed research-decision interface} (Sections~\ref{sec:taxonomy} and~\ref{sec:rdp}) is defined through a taxonomy of 24 types in six categories and an RDP connecting state, decisions, actions and evidence.
\item \textbf{A shared compact predictor} (Section~\ref{sec:candidate-token-prediction}) produces a typed answer and native option probabilities through one candidate-token readout in a single forward pass, supporting prediction-set construction from the same output distribution.
\item \textbf{A decision-level benchmark and empirical validation} (Section~\ref{sec:evaluation}) are provided through RD-Bench, which instantiates ten types in 2,000 rule-generated cases. DCRM (state-aware) raises ten-type accuracy from 27.8\% to 89.4\% over DCRM (question-only) and reduces mean prediction-set cardinality by 57.4\% at similar empirical coverage.
\end{enumerate}

\Needspace{8\baselineskip}
\section{Background and Related Work}
\label{sec:related-work}

\subsection{LLM Agents and Decision-Making}

LLM agents used reasoning, feedback and memory to guide action selection. ReAct interleaved reasoning traces with actions, allowing the model to reason before acting \cite{ref5}. Reflexion incorporated verbal self-feedback to support revisions in subsequent attempts \cite{ref6}. Chain-of-thought prompting elicited intermediate reasoning steps before an answer \cite{ref7}. Cognitive architectures for language agents organized such behavior into memory, planning and action modules \cite{ref8}, while later systems enabled agents to learn from accumulated experience \cite{ref9}. Surveys of LLM-based agents \cite{ref10,ref11} documented the rapid growth of this area.

These approaches address how agents reason, retain feedback and act. Assessing an individual decision raises a separate question about what was chosen and whether that choice was justified by the available evidence. When a choice is embedded in a generated reasoning trace, its alternatives and evaluation criterion must be identified before its quality can be assessed independently.

\subsection{Calibration, Prediction Sets and Cost-Aware Routing}

Calibration and selective prediction provide the foundations for interpreting decision confidence. Neural networks are often overconfident, and temperature scaling provides a simple and effective remedy \cite{ref12}. Prior studies estimated whether language models knew an answer through a dedicated value head trained on hidden states \cite{ref13} or by directly eliciting a probability \cite{ref14}. For models tuned with human feedback, the latter study found that elicited probabilities were often better calibrated than likelihood-based confidence. The broader uncertainty-quantification literature covered Bayesian, ensemble and distance-aware approaches \cite{ref27}, while LLM-specific surveys catalogued available estimators \cite{ref28,ref29}. Sampling-based estimators also provided black-box uncertainty estimates when logits were unavailable \cite{ref30}.

Conformal prediction converts an uncertainty score into a prediction set with a distribution-free coverage guarantee under exchangeability \cite{ref15}. Prior work used this framework to trigger human help for uncertain robot planners \cite{ref16} and extended it to open-ended generation from black-box LLMs \cite{ref17}. Recent work covered reject-option learning \cite{ref19}, selective prediction for natural language processing (NLP) \cite{ref20} and abstention in LLMs \cite{ref21}. A related framework, \textit{learning to defer}, learned a classifier and a rejector to decide when to defer to a downstream expert \cite{mozannar2020defer}. Cost-aware systems allocated computation through model cascades in FrugalGPT \cite{ref23}, cascades over reasoning representations \cite{ref24}, multi-level serving \cite{ref25} and token-budget-aware reasoning \cite{ref26}.

Applying these methods to scientific decisions requires confidence to have a clear referent, such as the correctness of a selected action or the inclusion of an appropriate alternative in a prediction set. Routing also depends on how uncertainty is weighed against the cost of further computation or human review. The problem is therefore to connect confidence over available choices to an explicit criterion for acting, seeking help or gathering more evidence.

\subsection{AI for Scientific Research}

AI has been applied across the research process, from hypothesis generation to experimental design and literature synthesis, and several surveys mapped this space \cite{ref31,ref32}. Recent multi-agent architectures organized debate and evolution over candidate hypotheses and validated them experimentally \cite{ref33}, while other systems automated the experimental loop \cite{ref34}. Benchmarks such as MLAgentBench \cite{ref35} and MLE-bench \cite{ref36} evaluated end-to-end machine-learning experimentation, while related efforts targeted information gathering \cite{ref37}, insight rediscovery \cite{ref38} and evidence-grounded evaluation \cite{ref39}. Agent evaluation methodology was also surveyed \cite{ref40}. Research State proposed a semantic abstraction for persistent scientific context \cite{ref41}, while a systematization of autonomous research identified challenges across systems, benchmarks and verification \cite{ref42}.

End-to-end outcomes reflect a sequence of choices together with the quality of tool use and execution. An unsuccessful run can arise from different points in that sequence, so the final result alone does not show which choices were appropriate. Evaluating research decisions therefore requires identifying the context in which each choice was made and the criterion by which that choice should be judged.

\section{Method}
\label{sec:method}

\subsection{Research Decision Taxonomy}
\label{sec:taxonomy}

A decision-centric account of research requires a definition of the decisions to be modeled. A research decision is defined as a recurring commitment that (i) can be posed as a question over the research state, (ii) admits a typed answer, and (iii) has a defensible ground truth against which the answer can be scored.

The decision interface supports recurring choices across research stages, so the same type can recur in different stages and several types can apply within one stage. Each type specifies its input fields, output space and evaluation metric, making its outputs independently measurable. These decisions connect to actions with different resource requirements, including tool calls, experimental runs and human review. To support downstream handling of uncertainty, each type exposes a confidence estimate whose relationship to correctness can be evaluated, and the interface permits a policy to send an uncertain decision to a stronger model or a person.

We define 24 research decision types organized into six categories, as shown in Table~\ref{tab:md-1}. Parenthesized answer spaces identify the ten types instantiated in RD-Bench. The category ordering reflects how state understanding informs evidence assessment, planning and agent control. Strategy operates at a higher level of commitment, while escalation governs when another decision-maker should participate. These relationships allow types to recur throughout a research project. Italicized names in the text identify decision types in this taxonomy. The identifiers in Table~\ref{tab:md-1} connect these names to the benchmark data and code.

\begin{table*}[!t]
\centering
\caption{Research decision taxonomy with six categories and 24 types, organized by their guiding question.}
\label{tab:md-1}
\begin{tabular}{>{\raggedright\arraybackslash}p{0.17\textwidth} >{\raggedright\arraybackslash}p{0.20\textwidth} >{\raggedright\arraybackslash}p{0.53\textwidth}}
\toprule
Category & Guiding question & Decision types \\
\midrule
A. State understanding & What is the situation? & A1~Research stage (7-way~label), A2~State completeness (4~levels), A3~Primary objective, A4~Open-problem ranking \\
B. Evidence and validity & Is the evidence adequate and sound? & B1~Evidence sufficiency (3~levels), B2~Claim-evidence consistency (4~levels), B3~Experiment validity, B4~Result anomaly, B5~Reproducibility risk \\
C. Planning and next action & What should happen next? & C1~Next action (4~candidates), C2~Action priority, C3~Plan continuation vs. revision, C4~Stopping condition (3~levels) \\
D. Research strategy & Should the direction change? & D1~Direction decision (4-way~label), D2~Hypothesis revision, D3~Scope change, D4~Methodology paradigm \\
E. Agent control & How should the agent operate? & E1~Tool selection, E2~Model routing (3~tiers), E3~Self-verification need, E4~Agent continuation vs. termination \\
F. Human and advisor escalation & When should a person be involved? & F1~Advisor review need (3~levels), F2~User confirmation need (binary), F3~Expert consultation trigger \\
\bottomrule
\end{tabular}
\par\smallskip
\begin{minipage}{0.97\textwidth}
\footnotesize\raggedright
Parentheses show benchmark answer spaces for the ten evaluated types. Each Next action case presents four candidates drawn from a ten-action vocabulary. Direction decision uses the labels continue, modify, branch and abandon.
\end{minipage}
\end{table*}

The taxonomy extends the decision interface across evidence, planning, strategy and escalation as well as tool selection. A typed output makes each choice independently scoreable, while a shared state representation connects choices that depend on the same evidence.

\subsection{Research Decision Process}
\label{sec:rdp}

The \textbf{Research Decision Process (RDP)} connects a research state $S_t=(Q_t,H_t,M_t,X_t,E_t,C_t,R_t,\tau_t)$ to a typed decision, an action and the evidence that updates the state. Its components represent the question, hypotheses, method, experiments, evidence, claims, resources and history. A decision predictor supplies $D_t$ for a selected type $d_k$. An action executor then realizes that choice, and the resulting evidence is used to update $S_{t+1}$.

\begin{equation}
\label{eq:rdp-transition}
S_t \xrightarrow{\;D_t=f_\theta(S_t,d_k)\;} A_t \xrightarrow{\;E_t\;} S_{t+1}.
\end{equation}

The interface in~\eqref{eq:rdp-transition} separates prediction, execution and state update. Each type defines its candidate space, while DCRM shares its language model and candidate-token readout across types. The RDP policy loop includes confidence-based escalation, action execution and evidence-driven state updates, as specified in Algorithm~\ref{alg:rdp}. Its progress measure connects the sequence of typed decisions to a cost-aware research objective.

\par\addvspace{8pt}
\noindent\begin{minipage}{\linewidth}
\small
\hrule height 0.6pt
\vspace{4pt}
\refstepcounter{rdpalgorithm}\label{alg:rdp}\textbf{Algorithm \therdpalgorithm}\quad Research Decision Process (RDP)
\par\vspace{4pt}
\hrule height 0.4pt
\renewcommand{\algorithmicrequire}{\textbf{Input:}}
\renewcommand{\algorithmicensure}{\textbf{Output:}}
\begin{algorithmic}[1]
\Require initial state $S_0$, decision model $f_\theta$, action executor $\mathcal{A}$, progress measure $\Delta P$
\Ensure states $S_0,\dots,S_T$, accumulated progress $\sum_t \Delta P_t$
\State $t \leftarrow 0$
\While{the termination condition is not met}
    \State select decision type $d_k \in \mathcal{T}$ for state $S_t$
    \State $D_t \leftarrow f_\theta(S_t, d_k)$
    \If{$\text{conf}(D_t) < \theta_{\text{abstain}}$}
        \State $D_t \leftarrow \text{escalate}(S_t, d_k)$
    \EndIf
    \State $A_t \leftarrow \mathcal{A}(D_t)$
    \State $E_t \leftarrow$ evidence observed from $A_t$
    \State $\Delta P_t \leftarrow \text{progress}(S_t, E_t) - \lambda \cdot C(A_t)$
    \State $S_{t+1} \leftarrow \text{update}(S_t, D_t, A_t, E_t)$
    \State $t \leftarrow t + 1$
\EndWhile
\end{algorithmic}
\hrule height 0.6pt
\end{minipage}\par\vspace{8pt}

A downstream policy can score an action by expected progress net of cost using the utility in~\eqref{eq:cost-aware-utility}.
\begin{equation}
\label{eq:cost-aware-utility}
\begin{aligned}
U(a\mid S)&=\mathbb{E}[w^\top\Delta P(S,a)]-\lambda\bar C(a), \\
\bar C(a)&=\sum_j\lambda_j c_j(a).
\end{aligned}
\end{equation}
The cost components encode normalized token, compute, wall-clock and human-time requirements. The progress vector in~\eqref{eq:research-progress} has components for knowledge, evidence strength, internal consistency, uncertainty and potential.
\begin{equation}
\label{eq:research-progress}
\Delta P=(\Delta K,\Delta E,\Delta C,-\Delta U,\Delta V).
\end{equation}
Application-specific weights satisfy $w_i\geq0$ and $\sum_iw_i=1$.

The RDP framework defines evaluation targets at three levels. Decision-level evaluation scores a typed answer and its confidence given a research state using Top-1/Top-3 accuracy, expected calibration error (ECE), Brier score and risk--coverage. Agent-level evaluation considers decision--action trajectories formed through state and tool interactions, with task success, cost and invalid-action rate as metrics. Research-level evaluation assesses the final state and artefacts of a project under a given question and budget through progress, output quality and cost-efficiency. This hierarchy connects individual predictions to agent behavior and research outcomes.

\subsection{Shared Candidate-Token Prediction}
\label{sec:candidate-token-prediction}

\textbf{Shared option distribution.} DCRM reads the state, typed question and candidate list through a shared causal language model. A restricted softmax over the candidate-marker logits yields the option distribution. Its argmax selects the answer, and a held-out calibration threshold turns the same probabilities into a prediction set. The output space is closed by construction, and answer prediction requires one forward pass without autoregressive generation.

\textbf{Input and readout.} The renderer emits the question, hypotheses, method, experiments, evidence and claims, with an optional history list. The decision question and option list specify the type. Candidate markers are verified as stable single-token continuations, so the shared language-model output head supplies a directly comparable logit for each candidate. The dataset metadata field \texttt{research\_stage} is excluded from model inputs.

\textbf{Supervised training.} Low-Rank Adaptation (LoRA) adapters \cite{ref44} were fitted by minimizing cross-entropy on the ground-truth candidate marker while the base language model was kept frozen. The candidate prediction task was learned in one supervised phase. At inference, the selected-candidate confidence is $\max_y p(y\mid S,d_k)$, while all candidate probabilities remain available to the downstream interface.

\textbf{Calibrated prediction sets.} For a held-out calibration example, the score is defined in~\eqref{eq:calibration-score}.
\begin{equation}
\label{eq:calibration-score}
s(S,y)=1-p(y\mid S,d_k).
\end{equation}
At target miscoverage $\alpha$, the threshold is the \mbox{$\lceil(n+1)(1-\alpha)\rceil$-th} ordered calibration score, or infinity when that rank exceeds $n$. The prediction set contains candidates with scores at most the threshold. An empty set is enlarged with the argmax candidate. Calibration and test examples use the same unmodified candidate-probability scale. Split conformal coverage has its standard guarantee under exchangeability \cite{ref15}. Section~\ref{sec:prediction-sets} reports empirical coverage under the held-out research-domain split.

\FloatBarrier

\section{Evaluation}
\label{sec:evaluation}

The evaluation examines how accurately DCRM (state-aware) predicts typed decisions, how compactly its probabilities represent plausible answers and what operating point a 0.6B predictor provides. All decision-quality tables use case-indexed stored predictions, with input hashes and deterministic candidate-order tie handling recorded in the evaluation manifest. The analyses connect prediction quality to statistical reliability, input information, inference configurations and resource use.

\FloatBarrier

\subsection{Experimental Setup}
\label{sec:setup}

\begin{table*}[!t]
\centering
\footnotesize
\setlength{\tabcolsep}{3pt}
\caption{Decision accuracy and candidate ranking across ten types}
\label{tab:md-3}
\begin{tabular*}{\textwidth}{@{\extracolsep{\fill}}lrrrrrr@{}}
\toprule
& \multicolumn{2}{c}{\shortstack{DCRM\\(state-aware)}} & \multicolumn{2}{c}{\shortstack{DCRM\\(question-only)}} & \shortstack{DeepSeek-\\V4.1-Flash} & Gain \\
\cmidrule(lr){2-3}\cmidrule(lr){4-5}
Decision type & Top-1 & Top-3 & Top-1 & Top-3 & Top-1 & (pp) \\
\midrule
Research stage & \textbf{0.767} & 0.867 & 0.107 & 0.327 & 0.260 & +66.0 \\
State completeness & \textbf{0.687} & 0.947 & 0.187 & 0.820 & 0.440 & +50.0 \\
Evidence sufficiency & \textbf{1.000} & 1.000$^{\dagger}$ & 0.353 & 1.000$^{\dagger}$ & 0.340 & +64.7 \\
Claim-evidence consistency & \textbf{0.827} & 0.967 & 0.227 & 0.793 & 0.240 & +60.0 \\
Next action & \textbf{0.720} & 0.933 & 0.273 & 0.720 & 0.340 & +44.7 \\
Stopping condition & \textbf{1.000} & 1.000$^{\dagger}$ & 0.353 & 1.000$^{\dagger}$ & 0.660 & +64.7 \\
Direction decision & \textbf{0.967} & 1.000 & 0.267 & 0.787 & 0.260 & +70.0 \\
Model routing & \textbf{0.973} & 1.000$^{\dagger}$ & 0.247 & 1.000$^{\dagger}$ & 0.300 & +72.7 \\
Advisor review need & \textbf{1.000} & 1.000$^{\dagger}$ & 0.267 & 1.000$^{\dagger}$ & 0.300 & +73.3 \\
User confirmation need & \textbf{1.000} & 1.000$^{\dagger}$ & 0.500 & 1.000$^{\dagger}$ & 0.500 & +50.0 \\
\midrule
All ten types & \textbf{0.894} & 0.971 & 0.278 & 0.845 & 0.364 & \textbf{+61.6} \\
\bottomrule
\end{tabular*}
\par\smallskip
\begin{minipage}{\textwidth}
\footnotesize\raggedright
Each type has 50 held-out test cases. Results for DCRM (state-aware) and DCRM (question-only) are means over three training seeds. DeepSeek-V4.1-Flash was evaluated once on the same 500 test cases and returned a single answer per case. Gain is the Top-1 difference between DCRM (state-aware) and DCRM (question-only) in percentage points (pp). Bold accuracies identify the highest Top-1 result in each row. $\dagger$ Top-3 includes all candidates for types with at most three options and therefore equals 1.000.
\end{minipage}
\end{table*}

\textbf{Benchmark.} RD-Bench contains 2,000 rule-generated cases spanning ten decision types and all six taxonomy categories. Each type has 200 cases across four research domains, including federated learning, graph representation learning, protein structure prediction and catalyst discovery. Domains f1/f2 supply 750 training and 250 calibration cases, while f3 supplies 500 validation cases and f4 supplies 500 test cases. Each case includes a Research State, a randomized candidate order and a programmatic answer. The benchmark measures agreement with explicit decision rules under the held-out-domain split.

\textbf{Comparisons.} DCRM (state-aware) and DeepSeek-V4.1-Flash receive the rendered Research State, typed decision question and candidates. DCRM (question-only) receives the research question, typed decision question and candidates during training and evaluation. The two DCRM variants are trained separately with the same base model, candidate-token readout, supervision targets and training recipe. Their comparison measures the contribution of decision-relevant state information. Top-1 exact match is common to all three models.

\textbf{Training.} DCRM (state-aware) was trained from Qwen3-0.6B-Base \cite{ref45} with LoRA rank 16, five training epochs, learning rate 5e-5, gradient accumulation 8 and maximum sequence length 2048. DCRM (question-only) was trained separately with the same recipe. Both models were trained with seeds 42, 43 and 44, while DeepSeek-V4.1-Flash was evaluated in one API evaluation run. Training and local measurement were performed on an RTX 4090 graphics processing unit (GPU) with 24 GB of memory, Ubuntu 20.04, Python 3.12.11, CUDA 12.6, PyTorch 2.7.1 and Transformers 4.53.2. Section~\ref{sec:runtime} specifies the timing and accounting procedures.

\textbf{API configuration.} DeepSeek-V4.1-Flash was evaluated through DeepSeek's chat-completions API using request name \texttt{deepseek-flash}.\footnote{The request name resolved to DeepSeek-V4.1-Flash during measurements from 2026-09-22 to 2026-09-28.} The evaluation covered 500 test cases with an initial output budget of 16,000 tokens. Each prompt supplied the Research State, typed question and candidates and requested one option letter, parsed as the last standalone candidate letter. Recorded responses averaged 1,372 reasoning tokens, identifying a reasoning-mode configuration.

\textbf{Benchmark construction and label sources.} Programmatic checks pass for all 2,000 cases. Full states are unique, and the rendered state text contains 1,855 distinct values with disjoint training and test text. The programmatic rules supply the supervised decision labels. A separate label-source diagnostic used 60 uniformly sampled cases from the earlier three-type benchmark. An independent blind model response matched the programmatic label in 15 cases (0.250), compared with a random reference of 0.230. These counts describe agreement between the two label sources.

\FloatBarrier

\subsection{Decision Accuracy and Candidate Ranking}
\label{sec:quality}

\begin{figure*}[t]
\centering
\includegraphics[width=0.98\textwidth]{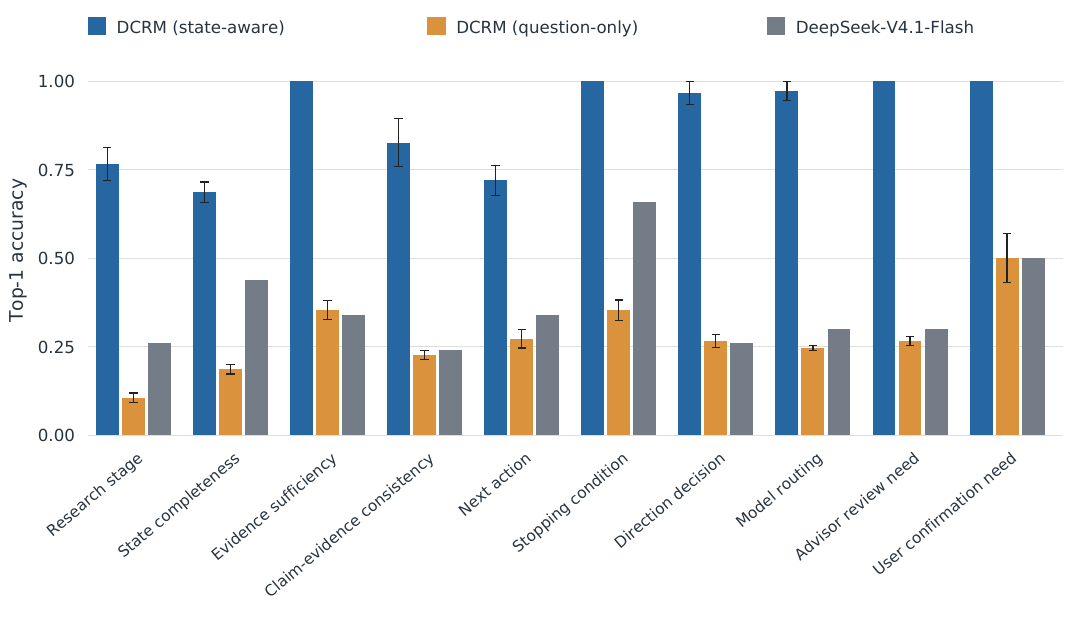}
\caption{Decision accuracy across the complete ten-type suite. DCRM (state-aware) and DCRM (question-only) share their base model and training recipe. Error bars show the standard error over three training seeds, whereas DeepSeek-V4.1-Flash was evaluated in one API evaluation run on the same cases. See Table~\ref{tab:md-1} for type definitions and benchmark answer spaces.}
\label{fig:decision-quality}
\end{figure*}

\textbf{State-aware supervision improves accuracy.} Across the complete ten-type suite, Top-1 accuracy rises from \textbf{27.8\%} for DCRM (question-only) to \textbf{89.4\%} for DCRM (state-aware), a gain of \textbf{61.6 percentage points}. Its advantage over DeepSeek-V4.1-Flash (\textbf{36.4\%}) is \textbf{53.0 percentage points}, as shown in Table~\ref{tab:md-3}. DCRM (state-aware) has higher mean accuracy on every type and reaches 1.000 on \decisiontype{Evidence sufficiency}, \decisiontype{Stopping condition}, \decisiontype{Advisor review need} and \decisiontype{User confirmation need}. \decisiontype{Direction decision} and \decisiontype{Model routing} reach 0.967 and 0.973, respectively. These gains extend across the complete per-type comparison shown in Figure~\ref{fig:decision-quality}.

\textbf{Candidate ranking and subset analysis.} Top-3 accuracy measures candidate ranking from the option probabilities, with ties resolved by recorded candidate order. For the complete ten-type suite, it is 0.971 for DCRM (state-aware) and 0.845 for DCRM (question-only), as reported in Table~\ref{tab:md-3}. A post-hoc analysis considered the six types on which the mean Top-1 accuracy of DCRM (state-aware) was below 1.000. On this subset of 300 test cases, Top-1 accuracy was 0.823 for DCRM (state-aware), 0.218 for DCRM (question-only) and 0.307 for DeepSeek-V4.1-Flash. Top-3 accuracy was 0.952 for DCRM (state-aware) and 0.741 for DCRM (question-only).

\textbf{Stage input and sensitivity.} The \decisiontype{Research stage} task uses the rendered state fields specified in Section~\ref{sec:candidate-token-prediction}. The dataset metadata field \texttt{research\_stage} matches the label for \decisiontype{Research stage} in 200/200 cases and remains outside the rendered prompt. Excluding \decisiontype{Research stage} from the six-type analysis yields 0.835 for DCRM (state-aware), 0.240 for DCRM (question-only) and 0.316 for DeepSeek-V4.1-Flash.

\FloatBarrier

\subsection{Paired Statistical Analysis}
\label{sec:statistics}

\begin{table*}[!htbp]
\centering
\footnotesize
\setlength{\tabcolsep}{3pt}
\caption{Holm-adjusted McNemar p-values for state-aware DCRM comparisons}
\label{tab:md-9}
\begin{tabular*}{\textwidth}{@{\extracolsep{\fill}}>{\raggedright\arraybackslash}p{96pt}rrrrrr@{}}
\toprule
& \multicolumn{3}{c}{\shortstack{DCRM (state-aware)\\vs DCRM (question-only)}} & \multicolumn{3}{c}{\shortstack{DCRM (state-aware)\\vs DeepSeek-V4.1-Flash}} \\
\cmidrule(lr){2-4}\cmidrule(lr){5-7}
& \multicolumn{3}{c}{DCRM training seed} & \multicolumn{3}{c}{DCRM training seed} \\
Decision type & 42 & 43 & 44 & 42 & 43 & 44 \\
\midrule
Research stage & $6.93\times10^{-7}$ & $1.49\times10^{-9}$ & $4.19\times10^{-9}$ & $6.39\times10^{-3}$ & $5.01\times10^{-6}$ & $1.42\times10^{-4}$ \\
State completeness & $1.18\times10^{-6}$ & $6.44\times10^{-6}$ & $3.71\times10^{-5}$ & $\mathbf{0.0541}$ & $\mathbf{0.115}$ & $\mathbf{0.132}$ \\
Evidence sufficiency & $1.49\times10^{-8}$ & $1.34\times10^{-9}$ & $1.49\times10^{-8}$ & $5.12\times10^{-9}$ & $5.12\times10^{-9}$ & $5.12\times10^{-9}$ \\
Claim-evidence consistency & $3.71\times10^{-5}$ & $7.57\times10^{-10}$ & $3.71\times10^{-5}$ & $2.53\times10^{-7}$ & $8.15\times10^{-10}$ & $9.54\times10^{-7}$ \\
Next action & $2.77\times10^{-4}$ & $7.59\times10^{-6}$ & $1.76\times10^{-4}$ & $\mathbf{0.0541}$ & $8.18\times10^{-4}$ & $1.97\times10^{-2}$ \\
Stopping condition & $2.42\times10^{-8}$ & $1.34\times10^{-9}$ & $7.92\times10^{-9}$ & $1.53\times10^{-4}$ & $1.53\times10^{-4}$ & $1.53\times10^{-4}$ \\
Direction decision & $2.18\times10^{-10}$ & $1.34\times10^{-9}$ & $6.93\times10^{-7}$ & $4.37\times10^{-10}$ & $4.37\times10^{-10}$ & $8.85\times10^{-9}$ \\
Model routing & $2.21\times10^{-9}$ & $3.93\times10^{-10}$ & $2.18\times10^{-10}$ & $1.68\times10^{-8}$ & $1.57\times10^{-9}$ & $1.57\times10^{-9}$ \\
Advisor review need & $7.57\times10^{-10}$ & $2.18\times10^{-10}$ & $7.57\times10^{-10}$ & $1.57\times10^{-9}$ & $1.57\times10^{-9}$ & $1.57\times10^{-9}$ \\
User confirmation need & $6.93\times10^{-7}$ & $2.29\times10^{-5}$ & $1.49\times10^{-8}$ & $9.54\times10^{-7}$ & $9.54\times10^{-7}$ & $9.54\times10^{-7}$ \\
\bottomrule
\end{tabular*}
\par\smallskip
\begin{minipage}{\linewidth}
\footnotesize\raggedright
Entries are Holm-adjusted two-sided exact McNemar p-values for the comparisons named above. \textbf{Bold values} indicate $p\geq0.05$ and do not reach significance at $\alpha=0.05$. Each test pairs the two models on the same 50 cases within a decision type. Holm correction is applied separately to each comparator's family of 30 tests (ten types and three training seeds). Seeds identify DCRM (state-aware) training runs, with matching seeds for DCRM (question-only). DeepSeek-V4.1-Flash was evaluated once on the test set, and the same predictions are reused for all three seeds.
\end{minipage}
\end{table*}

McNemar's exact test uses the discordant pairs on identical test cases. Paired bootstrap intervals were computed from 10,000 resamples with random seed 20261019. Each training seed defines a paired comparison on the shared case set. At significance level $\alpha$=0.05, Holm correction is applied separately to the ten-type DCRM (question-only) family and the ten-type DeepSeek-V4.1-Flash family, each containing 30 type--seed tests. Training seeds reuse test cases, and the paired unit is the case within a seed.

Table~\ref{tab:md-9} reports the adjusted p-values for all ten types against both comparators. All 30 comparisons against DCRM (question-only) pass Holm correction. Against DeepSeek-V4.1-Flash, DCRM (state-aware) has higher observed Top-1 in all 30 paired comparisons, with 28 passing the unadjusted 0.05 threshold and 26 passing Holm correction. The adjusted values are 0.0541 for \decisiontype{State completeness} and \decisiontype{Next action} at seed 42, and exceed 0.05 for \decisiontype{State completeness} at seeds 43 and 44.

On the six-type subset, the pooled accuracy difference for DCRM (state-aware) across the three seeds is +0.573 to +0.663 over DCRM (question-only) and +0.487 to +0.573 over DeepSeek-V4.1-Flash, as reported in Table~\ref{tab:md-10}. The paired 95\% confidence intervals are positive for both comparisons, and all pooled exact McNemar p-values are below 0.0001. These intervals complement the per-type tests by quantifying the magnitude of the pooled improvements.

\begin{table}[!htbp]
\centering
\footnotesize
\setlength{\tabcolsep}{3pt}
\caption{Paired accuracy differences and confidence intervals}
\label{tab:md-10}
\begin{tabular}{@{}l r r r r@{}}
\toprule
& \multicolumn{2}{c}{\shortstack{DCRM\\(question-only)}} & \multicolumn{2}{c}{DeepSeek-V4.1-Flash} \\
\cmidrule(lr){2-3}\cmidrule(lr){4-5}
Seed & Diff. & 95\% CI & Diff. & 95\% CI \\
\midrule
42 & +0.580 & [0.517, 0.643] & +0.490 & [0.413, 0.563] \\
43 & +0.663 & [0.603, 0.720] & +0.573 & [0.500, 0.640] \\
44 & +0.573 & [0.507, 0.637] & +0.487 & [0.410, 0.560] \\
\bottomrule
\end{tabular}
\par\smallskip
\begin{minipage}{\linewidth}
\footnotesize\raggedright
Each seed uses the same 300 cases in the six-type subset. Differences are DCRM (state-aware) minus the named comparison model on the 0--1 accuracy scale. Intervals are paired bootstrap 95\% confidence intervals (CIs). All pooled exact McNemar p-values are below 0.0001.
\end{minipage}
\end{table}

\subsection{Prediction Sets and Confidence Quality}
\label{sec:prediction-sets}

\textbf{Compact alternatives at high empirical coverage.} A decision layer can provide a prediction set alongside its top answer. Per-type split-conformal prediction sets were evaluated at nominal coverage 0.95 on the ten-type benchmark. The 250 calibration examples are distributed across types in counts of 20--29, and each threshold uses its actual type-specific sample count. Calibration uses domains f1/f2, whereas testing uses f4. Coverage is the fraction of ground-truth labels included in the predicted sets. Both models use the same raw-probability nonconformity score and finite-sample quantile.

\begin{figure*}[t]
\centering
\includegraphics[width=\textwidth]{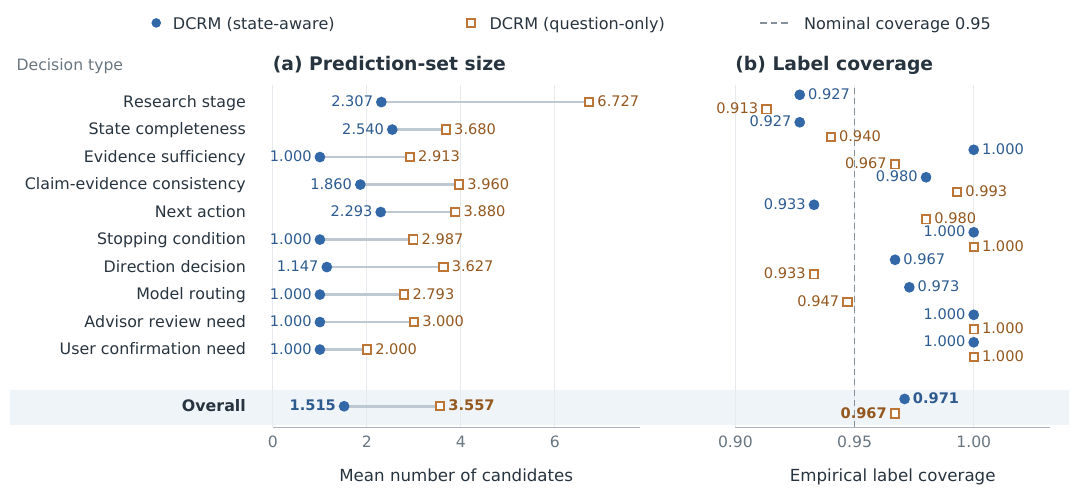}
\caption{DCRM (state-aware) produces smaller prediction sets than DCRM (question-only) at similar overall empirical label coverage. (a) Mean set cardinality by decision type. (b) Empirical ground-truth label inclusion, with the dashed line marking nominal coverage 0.95. Each type has 50 test cases, and metrics average three training seeds. The highlighted Overall results average all ten types using unrounded values.}
\label{fig:prediction-set-efficiency}
\end{figure*}

Figure~\ref{fig:prediction-set-efficiency} compares prediction-set size and empirical label coverage by decision type and overall. Across the suite, DCRM (state-aware) returns \textbf{1.515 candidates at 0.971 empirical coverage}, compared with 3.557 at 0.967 for DCRM (question-only). At similar achieved label coverage, this is a \textbf{57.4\% reduction in mean set cardinality}. DCRM (state-aware) produces smaller mean sets on every type in Figure~\ref{fig:prediction-set-efficiency}(a), with corresponding empirical coverages shown against the nominal target in panel (b). After threshold calibration, the distribution therefore supports a compact prediction set, allowing a downstream verifier or advisor to inspect the selected answer together with a small set of alternatives.

\textbf{Probability quality and correctness ranking.} The candidate probabilities can also be assessed through confidence quality and correctness ranking. P(IK) denotes the probability that a model knows the answer, where IK stands for ``I know'' \cite{ref13}. Inspired by this approach, a post-hoc logistic regression probe was fitted on final-layer hidden states using hard binary correctness labels from the calibration split, with the language model frozen. We compared this probe with native selected-candidate probability on the 1.7B model at seed 42, with 50 test cases per type, as reported in Table~\ref{tab:md-13}. ECE compares confidence with binary correctness in ten equal-frequency bins, Brier measures probability quality and area under the receiver operating characteristic curve (AUROC) measures correctness ranking.

\begin{table}[!htbp]
\centering
\footnotesize
\setlength{\tabcolsep}{2pt}
\caption{Confidence quality and correctness ranking}
\label{tab:md-13}
\begin{tabular}{@{}>{\raggedright\arraybackslash}p{64pt} l r r r r@{}}
\toprule
Decision type & Signal & Accuracy & ECE & Brier & AUROC \\
\midrule
Evidence sufficiency & softmax & 0.860 & 0.055 & 0.089 & 0.882 \\
Evidence sufficiency & P(IK) & 0.860 & 0.153 & 0.159 & 0.791 \\
\addlinespace[3pt]
Next action & softmax & 0.600 & 0.226 & 0.286 & 0.542 \\
Next action & P(IK) & 0.600 & 0.238 & 0.295 & 0.638 \\
\addlinespace[3pt]
Research stage & softmax & 0.300 & 0.371 & 0.325 & 0.743 \\
Research stage & P(IK) & 0.300 & 0.339 & 0.332 & 0.585 \\
\bottomrule
\end{tabular}
\par\smallskip
\begin{minipage}{\linewidth}
\footnotesize\raggedright
Results use the exploratory three-type benchmark, the 1.7B model and seed 42, with 50 test cases per type. Lower ECE and Brier scores and higher accuracy and AUROC are preferred.
\end{minipage}
\end{table}

Native candidate confidence has lower Brier scores on all three types. The logistic regression probe improves \decisiontype{Next action} correctness ranking from AUROC 0.542 to 0.638, whereas native confidence gives higher AUROC on \decisiontype{Evidence sufficiency} and \decisiontype{Research stage}. These measurements describe distinct probability-quality and ranking properties of the two signals.

\FloatBarrier

\subsection{State Inputs and Task Adaptation}
\label{sec:ablation}

The exploratory three-type benchmark covers \decisiontype{Research stage}, \decisiontype{Evidence sufficiency} and \decisiontype{Next action}, with 50 test cases per type. To assess the contributions of the current state and history to typed prediction, we compared DCRM (state-aware), its history-ablated variant and DCRM (question-only) at 0.6B over three training seeds, as summarized in Table~\ref{tab:md-5}.

\begin{table}[!htbp]
\centering
\footnotesize
\setlength{\tabcolsep}{3pt}
\caption{State-input ablation on three decision types}
\label{tab:md-5}
\begin{tabular}{@{}>{\raggedright\arraybackslash}p{103pt} rrr@{}}
\toprule
Variant & \shortstack{Research\\stage} & \shortstack{Evidence\\sufficiency} & \shortstack{Next\\action} \\
\midrule
DCRM (state-aware) & 0.347 & 0.760 & 0.633 \\
History-ablated DCRM & 0.373 & 0.727 & 0.593 \\
DCRM (question-only) & 0.247 & 0.447 & 0.440 \\
\bottomrule
\end{tabular}
\par\smallskip
\begin{minipage}{\linewidth}
\footnotesize\raggedright
Mean Top-1 is reported over three training seeds with 50 test cases per type.
\end{minipage}
\end{table}

\textbf{State fields deliver the main input gain.} Research State input raises accuracy from 24.7\% to 34.7\% for \decisiontype{Research stage}, 44.7\% to 76.0\% for \decisiontype{Evidence sufficiency} and 44.0\% to 63.3\% for \decisiontype{Next action}, as shown in Table~\ref{tab:md-5}. The corresponding gains over question-only input are 10.0, 31.3 and 19.3 percentage points. History removal changes accuracy by +2.7, -3.3 and -4.0 percentage points, respectively. These gains support prioritizing decision-relevant state fields in the compact predictor, while history effects vary by type.

\textbf{Task adaptation at a fixed backbone.} On the same 150 held-out cases spanning \decisiontype{Research stage}, \decisiontype{Evidence sufficiency} and \decisiontype{Next action}, LoRA adaptation raises Qwen3-4B Top-1 accuracy from 18.7\% (28/150) to 57.3\% (86/150), a gain of \textbf{38.7 percentage points}. Both models receive the full rendered state and use the same candidate-token readout. The adapted model uses training seed 42 and supplies the 4B operating point analyzed in Section~\ref{sec:runtime}. This comparison supports task-specific supervision for the state-conditioned decision interface.

\FloatBarrier

\subsection{Inference Configuration Analysis}
\label{sec:configurations}

The application programming interface (API) configuration analysis examines how response format and repeated API runs relate to typed decision accuracy.

\begin{table*}[!t]
\centering
\footnotesize
\setlength{\tabcolsep}{3pt}
\caption{Decision quality and runtime across inference configurations}
\label{tab:md-12}
\begin{tabular*}{\textwidth}{@{\extracolsep{\fill}}llrrrr@{}}
\toprule
Model & Format & Top-1 & \shortstack{API end-to-end\\(s)} & \shortstack{Local GPU forward\\(ms)} & \shortstack{Completion\\tokens} \\
\midrule
DeepSeek-V4.1-Flash, 2 runs & plain & 0.347 / 0.327 & 4.4--4.9 & --- & 790--905 \\
DeepSeek-V4.1-Flash & structured & 0.340 & 3.9 & --- & 700 \\
DeepSeek-V4.1-Flash & plain & 0.320 & 5.2 & --- & 870 \\
\textbf{DCRM (state-aware), 0.6B} & structured state & \textbf{0.580} & --- & 13.1 & --- \\
\bottomrule
\end{tabular*}
\par\smallskip
\begin{minipage}{\linewidth}
\footnotesize\raggedright
Each run uses the exploratory three-type benchmark of 150 cases. The two additional DeepSeek-V4.1-Flash runs report individual accuracies, with timing and token counts shown as ranges. API time includes end-to-end serving and generation, while local time covers GPU forward execution. DCRM (state-aware) uses a single forward pass. Dashes denote entries without the corresponding measurement or generation statistic.
\end{minipage}
\end{table*}

Table~\ref{tab:md-12} reports response-format and repeated-run diagnostics on the exploratory three-type benchmark, with 150 cases per run. Across the tested plain and structured formats, DeepSeek-V4.1-Flash accuracy is 0.320--0.340. Two additional DeepSeek-V4.1-Flash runs supplement the plain-format evaluation. The ten-type DeepSeek-V4.1-Flash result in Table~\ref{tab:md-3} comes from a separate API evaluation of the 500 test cases.

DCRM (state-aware) achieves 0.580 Top-1 on this three-type benchmark through the shared candidate-token readout, with prediction quality and execution profiles for each response configuration reported in Table~\ref{tab:md-12}. The API configurations provide complementary operating points with generated responses, while the local predictor returns its option probabilities directly after one forward pass.

\subsection{Runtime and Component Accounting}
\label{sec:runtime}

The resource analysis connects model size, typed prediction quality and execution cost. We measured the operating points of 0.6B, 1.7B and 4B models on the exploratory three-type benchmark, as summarized in Table~\ref{tab:md-6}.

\begin{table}[!htbp]
\centering
\footnotesize
\setlength{\tabcolsep}{4pt}
\caption{Model size, accuracy and resource footprint}
\label{tab:md-6}
\begin{tabular}{lrrrr}
\toprule
\shortstack{Model\\size} & Top-1 & Seeds & \shortstack{Forward\\time (ms)} & \shortstack{Peak GPU\\memory (GiB)} \\
\midrule
0.6B & 0.580 & 3 & 13.1 & 1.3 \\
1.7B & 0.589 & 3 & 14.4 & 3.4 \\
4B & 0.573 & 1 & 26.0 & 7.7 \\
\bottomrule
\end{tabular}
\par\smallskip
\begin{minipage}{\linewidth}
\footnotesize\raggedright
All model sizes use DCRM (state-aware). Top-1 averages the listed training seeds on the exploratory three-type benchmark. Forward time and peak allocated GPU memory were measured with seed-42 adapters. Each timing run used 147 predictions at batch size one. The timer boundaries and device snapshots are specified in the text.
\end{minipage}
\end{table}

\textbf{A 1.3-GiB decision predictor.} Across the evaluated model sizes, observed Top-1 ranges from 0.573 to 0.589. The 0.6B model combines 0.580 accuracy with the smallest measured memory footprint, 1.3 GiB, and a recorded mean forward component of 13.1 ms. The 1.7B and 4B models use 3.4 and 7.7 GiB, respectively. These measurements identify a compact operating point for this benchmark, where a small shared predictor handles typed decisions with direct access to all candidate probabilities.

\textbf{Local timing.} Each size was measured using its seed-42 exploratory three-type adapter and 147 predictions with batch size one. The timer enclosed the GPU forward pass and candidate-logit extraction. Prompt rendering, tokenization and host-to-device transfer occurred before the timed region. Peak memory is the peak allocated GPU memory. The 0.6B run records a mean of 13.146 ms and a ninety-fifth-percentile (p95) latency of 13.651 ms. Device-utilization snapshots before the 0.6B, 1.7B and 4B measurements are 40\%, 2\% and 30\%, respectively.

\textbf{API timing.} Remote latency includes serving and generated completion tokens. DeepSeek-V4.1-Flash averages 5.443 seconds. For the exploratory runs, API latency is reported end to end, while local latency covers the GPU-forward component, as specified in Table~\ref{tab:md-12}.

\textbf{Fixed-coefficient accounting scenario.} We apply assumed coefficients of 0.35 USD per GPU-hour and 1.10 USD per million completion tokens to obtain the accounting estimates in Table~\ref{tab:md-15}. The local row prices the measured forward component, while the API row prices completion tokens. Input-token charges, training, idle capacity and serving overhead are excluded from this scenario. This component-based accounting connects each observed workload to its stated resource coefficient.

\begin{table}[!htbp]
\centering
\footnotesize
\setlength{\tabcolsep}{3pt}
\caption{Component costs under fixed accounting coefficients}
\label{tab:md-15}
\begin{tabular}{@{}>{\raggedright\arraybackslash}p{85pt} >{\raggedright\arraybackslash}p{85pt} r@{}}
\toprule
Model & Assumed unit price & \shortstack{Cost per decision\\($10^{-6}$ USD)} \\
\midrule
DCRM (state-aware), 0.6B & 0.35 USD/GPU-hour & 1.278 \\
DeepSeek-V4.1-Flash & 1.10 USD per million completion tokens & 1057 \\
\bottomrule
\end{tabular}
\end{table}

\FloatBarrier

\section{Discussion}
\label{sec:discussion}

\textbf{State-aware supervision makes decisions a distinct modeling target.} The 89.4\% Top-1 accuracy of DCRM (state-aware) exceeds the 27.8\% of DCRM (question-only) by 61.6 percentage points, showing the value of decision-relevant context for compact typed prediction on RD-Bench. The shared readout covers ten types spanning all six taxonomy categories. This aligns representation, supervision and evaluation around the individual choice and makes each decision directly inspectable.

\textbf{One option distribution supports answer selection and uncertainty inspection.} The probabilities of DCRM (state-aware) support both a top answer and, after calibration, prediction sets with 57.4\% fewer candidates than DCRM (question-only) at similar empirical coverage. This joint interface allows an agent to pass an uncertain choice to another component while preserving the alternatives considered by the decision layer through the candidate list and probabilities.

\textbf{Compact prediction provides a focused computational contract.} The implemented model shares a candidate-token readout across types and terminates answer prediction after one forward pass. The exploratory size comparison places 0.6B at a low-memory operating point. Together, typed supervision and direct probability access make the decision layer a concrete component that can be evaluated separately from the action executor and generator.

\textbf{A common contract for research policies.} The RDP connects typed answers and candidate probabilities to actions, evidence updates and progress objectives. Its three evaluation levels provide a common vocabulary for describing individual decisions, action sequences and project outputs. Combined with the shared predictor, this contract makes the decision layer an explicit part of a scientific-agent architecture.

\section{Conclusion}
\label{sec:conclusion}

Decision-Centric Research makes scientific choices explicit, typed and individually measurable. On ten rule-generated decision types, a 0.6B DCRM (state-aware) achieves \textbf{89.4\% Top-1 accuracy}, compared with 27.8\% for DCRM (question-only) and 36.4\% for DeepSeek-V4.1-Flash. After held-out calibration, its native option probabilities yield prediction sets containing \textbf{1.515 candidates at 97.1\% empirical label coverage}, reducing mean set size by \textbf{57.4\%} relative to DCRM (question-only) at similar coverage. The resulting design is a compact state-aware decision layer in which one shared predictor supplies an answer and an inspectable option distribution in a single forward pass.

\FloatBarrier

\bibliographystyle{IEEEtranN}
\bibliography{references}

\end{document}